\documentclass[sigconf,nonacm]{acmart}
\AtBeginDocument{%
  }
\usepackage{multirow}
\usepackage{subcaption}

\begin{document}

\title{MoRAX: Mobility-based Representation Augmentation for Geospatial Foundation Models}

\author{Ya Wen}
\affiliation{%
  \institution{The University of Hong Kong}
  \city{Hong Kong SAR}
  \country{China}
}
\email{wenya@connect.hku.hk}

\author{Jixuan Cai}
\affiliation{%
  \institution{The Chinese University of Hong Kong}
  \city{Hong Kong SAR}
  \country{China}
}
\email{codyjcai@gmail.com}

\author{Yulun Zhou}
\authornote{Corresponding authors.}
\affiliation{%
  \institution{The University of Hong Kong}
  \city{Hong Kong SAR}
  \country{China}
}
\email{yulunzhou@hku.hk}

\author{Alec Kirkley}
\authornotemark[1]
\affiliation{%
  \institution{The University of Hong Kong}
  \city{Hong Kong SAR}
  \country{China}
}
\email{akirkley@hku.hk}

\renewcommand{\shortauthors}{Ya Wen et al.}

\begin{abstract}
Geospatial Foundation Models (GFMs) are emerging as a powerful paradigm for learning semantically rich and geographically consistent visual and physical representations. However, their reliance on Earth-observation (EO) data leaves information about human activity largely underrepresented. Human mobility data reveals the functional and relational structure between regions that is missing from EO data, but is often limited only to the city where it is observed, making it challenging to use for transferable urban representation learning. We introduce \textbf{MoRAX}, a lightweight framework for augmenting geospatial embeddings with functional structure derived from human mobility. MoRAX preserves the coverage and consistency of a GFM while providing information about the functional connectivity among urban regions, permitting zero-shot deployment in unseen cities with or without available mobility data. Across four target cities spanning two countries, the MoRAX teacher model, which observes mobility, consistently outperforms GFMs and strong urban representation baselines in eight socioeconomic and environmental prediction tasks. Meanwhile, the student model, which never takes mobility data as input, approaches the teacher in performance on most tasks. Transfer results across countries further demonstrate that modulation conditioned on mobility flows provides a general mechanism for grounding geospatial foundations in the human dimension of cities.
\end{abstract}


\maketitle

\section{Introduction}

Urban region representation learning aims to encode spatial units of a city into low-dimensional vectors that support a wide range of downstream applications, including socioeconomic data estimation, environmental monitoring, and urban planning \cite{balsebre2024city, wen2024demo2vec,li2024urban, sun2026urbanverse}. A useful region representation should describe not only the physical and environmental properties of a space but also the functional role that the place plays within the broader urban system. Learning such representations is therefore an important step toward general-purpose urban intelligence \cite{chen2025self,zhang2024urban}.

Recent advances in geospatial foundation models (GFMs) provide a promising basis for this goal. By pretraining on large-scale Earth observation (EO) data, these models produce geographically extensive, rich representations of the natural and built environment~\cite{brown2025alphaearth,feng2026tessera}. AlphaEarth Foundations~\cite{brown2025alphaearth}, for instance, integrates multi-source temporal and spatially contextualized Earth observations into globally available, fine-grained geospatial embeddings. Meanwhile, remote-sensing foundation models such as RemoteCLIP~\cite{liu2024remoteclip} extract visual cues directly from satellite imagery. Compared with conventional heterogeneous urban data sources such as points of interest, land-use records, and demographic statistics, geospatial foundation representations offer more consistent, broader coverage across cities and are far less dependent on local taxonomies, individual data platforms, or city-specific collection processes~\cite{rolf2021generalizable,brown2025alphaearth}. However, by construction, GFMs learn semantics from what a region and its immediate surrounding environment physically look like, ignoring the fundamentally different (and equally important) information embedded in the \textit{relational} semantics of a city which are defined by how people actually move through, use, and connect urban regions~\cite{liu2025beyond}. Traditional GFMs thus remain incomplete for providing a holistic understanding of urban regions.

Human mobility, represented by observed flows of people between regions, has repeatedly been shown to reveal the functional and interaction-driven structure of cities that GFMs do not explicitly capture \cite{wen2026mora, HREP,yong2024musecl}. It is thus important to investigate how human mobility data can complement geospatial foundation representations. However, treating GFMs and human mobility data as symmetric modalities within a conventional fusion architecture is likely to underutilise their distinct roles and respective strengths. GFMs provide a compact representation of observable place characteristics and feature wide geographical coverage. Human mobility data instead captures relational dynamics among places. Moreover, since mobility data collection process requires dedicated infrastructure, it is often only available in a limited and potentially unrepresentative subset of cities~\cite{luca2021survey}.

This asymmetry motivates us to keep the GFM representation as a shared base and use mobility to adapt it through lightweight feature-wise modulation. Rather than reconstructing the GFMs' semantic space, the mobility signal changes how a region is represented according to its exchange relationships with the rest of the city. By design, the adaptation is a compact, self-contained module that can be learned separately from the base representation it acts upon. We therefore further distill the teacher's mobility-induced
modulation into a student model that does not depend on mobility data at inference time, but predicts similar modulation from widely available geospatial signals. This makes it possible to retain the broad geographic coverage of EO-based GFMs while introducing human-centric relational information. The resulting adaptation and distillation framework are applicable across different GFMs, such as AlphaEarth or RemoteCLIP without changing the overall framework.

\begin{figure*}
    \centering
    \includegraphics[
    width=0.6\linewidth,
    trim=90 5 140 5,
    clip
]{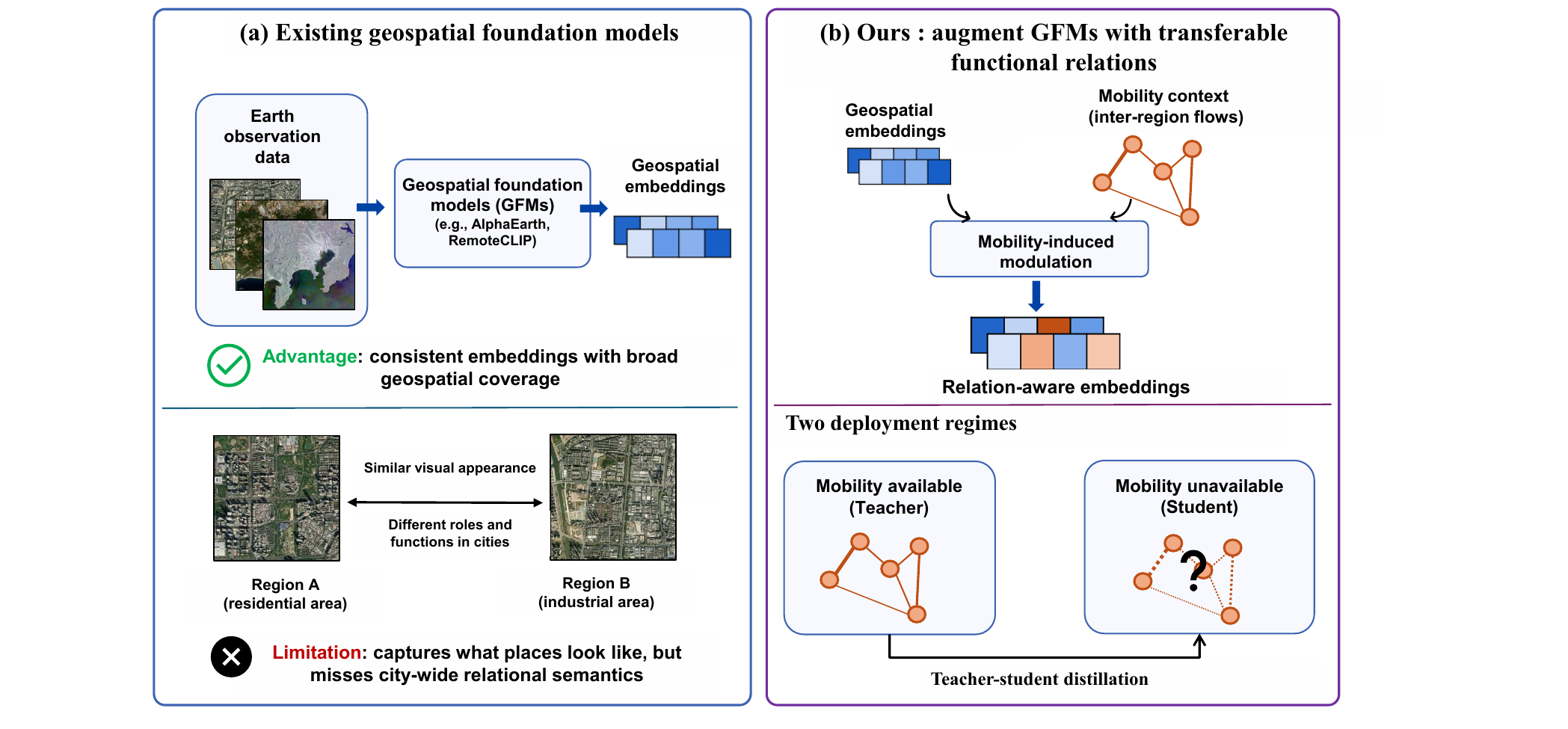}

    \caption{\textbf{Motivation and overview of MoRAX.} \normalfont Geospatial foundation models (GFMs) offer broad, consistent coverage but miss the relational semantics of how places function together (a). MoRAX adapts a frozen GFM base through lightweight modulation induced by mobility (b), supporting zero-shot deployment in unseen cities with or without available mobility data.}
    \label{fig:intro}
\end{figure*}

Based on this perspective, we propose \textbf{MoRAX}, a framework that learns how mobility adapts an existing geospatial foundation representation and distills this adaptation capability into a deployable, mobility-free encoder. A \textit{teacher} is trained on cities with observed mobility graphs, learning a lightweight modulation network that reshapes a baseline foundation embedding into a mobility-aware representation. A \textit{student}, trained on the same cities but never taking mobility as input---mobility influences it only through the teacher's supervision---is distilled to reproduce this modulation using a proxy relational signal constructed from widely available covariates alone. MoRAX consequently supports two complementary transfer regimes. When mobility is available in an unseen city, the pretrained teacher model directly produces mobility-aware region representations without the need to fine-tune the encoder on the target city. When mobility data is unavailable, the pretrained student model infers how mobility would have reshaped each region's representation, using only geospatial embeddings and pairwise geographic distances. Fig.~\ref{fig:intro} provides an overview of the MoRAX framework.

Our main contributions are summarized as follows:
\begin{itemize}
    \item We develop the MoRAX framework to augment EO-based GFMs with functional mobility structure through lightweight feature-wise modulation.
    \item We propose a teacher-student framework that distills this modulation into a student model, enabling functionally aware region representations in cities where no mobility data exists. 
    \item We test MoRAX in a series of zero-shot evaluation tasks, finding that the mobility-aware teacher consistently
    outperforms both geospatial foundation representations and urban
    representation learning baselines, while the mobility-free student remains competitive with the strongest baselines. 
    
\end{itemize}

\section{Related Work}

\textbf{Transferable Urban Region Representation Learning.}
Existing urban region representation methods draw on a wide range of data sources, including points of interest~\cite{huang2023learning,sun2024urban,gao2022geobert,yong2024musecl}, mobility flows~\cite{HREP,wu2022multi,Wang2017}, and satellite imagery~\cite{sun2025flexireg,klemmer2025satclip,Liu2023know}, which are typically integrated through multimodal fusion and alignment then trained and evaluated within the same city. This diversity in data sources makes fair comparison across methods difficult. At the same time, evaluating models within a single city makes it hard to tell whether they are learning patterns that generalize across urban systems or simply exploting correlations specific to one city or its data collection process. These representations thus offer limited capability of generalization to cities with different urban structure and data modality availability~\cite{sun2026urbanverse}.

Existing work has explored large scale urban representation pretraining by organizing heterogeneous urban signals
around mobility as the principal relational backbone. By learning from a nation-scale mobility graph, MoRA~\cite{wen2026mora} establishes a shared representation space in which
human-centered urban knowledge can be accumulated, compared, and reused
across locations. Similarly, PDFM~\cite{agarwal2024general} models
population dynamics from a rich collection of human-related signals
over a large scale neighborhood graph. These studies demonstrate the
potential of foundation-style pretraining for learning transferable
representations of urban systems from human activity and population
dynamics. However, their learned foundations tend to have more limited geographic coverage than EO foundation models (PDFM, for instance, covers only postal codes and counties within the United States).

\textbf{Earth Observation-Based Geospatial Foundation Models.} 
Benefiting from the growing volume, diversity, and geographic coverage
of Earth observation data, recent EO-based geospatial foundation models have
learned increasingly general and reusable representations of the
Earth's surface. For example, AlphaEarth Foundations~\cite{brown2025alphaearth}
integrates multi-source, temporal, and spatially contextualized
Earth observations into globally available geospatial embeddings while related efforts such as Tessera~\cite{feng2026tessera} similarly
seek to provide broadly applicable representations for downstream
geospatial applications. 
RemoteCLIP~\cite{liu2024remoteclip} aligns remote sensing imagery with
language to support transfer across classification, retrieval, and
recognition tasks. However, these representations remain primarily grounded in
EO signals and may not fully capture the
interaction-driven functional relationships that are critical to urban
region understanding.

\textbf{Adaptation of Pretrained Foundation Models.}
As pretrained foundation models have grown in scale and capability,
research across different domains has developed
a range of approaches for adapting their representations to new tasks,
domains, and information sources without retraining from scratch \cite{houlsby2019parameter,lester2021power,hu2022lora}. These methods parameterize adaptation through a compact set of task- or domain-specific trainable parameters, learned once and then held fixed at inference time. A complementary line of work instead performs adaptation dynamically, generating the adaptation itself as a function of an auxiliary conditioning signal rather than learning a static set of parameters. For example, FiLM~\cite{perez2018film} uses a conditioning signal to generate feature-wise scaling and shifting parameters that modulate intermediate activations. Similar ideas have been adopted in adaptive normalization and diffusion models, where representations are modulated according to external context such as style, class, text, or diffusion timestep ~\cite{huang2017adain,dhariwal2021diffusion}. Despite their broad use in language and vision, conditional adaptation
mechanisms remain underexplored in geospatial representation learning.
Geospatial foundation models are still typically used as fixed feature
extractors, with limited study of how complementary signals such as
city-wide relational information can adapt their pretrained
representations.

\section{Preliminaries}

\textbf{Regions.}
A city \(c\) is partitioned into \(N_c\) disjoint urban regions, denoted as
\(
\mathcal{R}^c=\{r_1^c,r_2^c,\ldots,r_{N_c}^c\}.
\)
For notational simplicity, we omit the city superscript \(c\) when the context is clear.

\textbf{Region Features.} Each region is associated with features drawn from different sources, characterizing it from complementary dimensions. We consider two data types, chosen for their independence from manually defined taxonomies (e.g., POI category schemes).

\textit{Definition 3.1 (Human Mobility Graph).}
Human mobility flows record movements between urban regions and reflect functional interactions among them. For a city with observed mobility data, we represent these interactions as a mobility graph
\(\mathcal{G}_m=(\mathcal{R},\mathcal{E}_m)\), where each edge
\((r_i,r_j)\in\mathcal{E}_m\) is associated with an observed interaction intensity
\(w_{ij}\) between regions \(r_i\) and \(r_j\). The intensity $w_{ij}$ is a generic measure of directed interaction between $r_i$ and $r_j$ within a given period of time, and provides a proxy for the functional connectivity of the two regions. It can be derived, for example, from observed flows of people or trips between the two regions, or from other interaction records aggregated to the region level.

\textit{Definition 3.2 (Geospatial Foundation Embedding).}
Each region $r_i$ is represented by a pretrained foundation embedding
$\mathbf{s}_i$, which summarizes the region's EO-derived physical,
environmental, and spatial characteristics in a globally
consistent representation space. Such embeddings are produced by encoders
pretrained on large-scale EO corpora, including satellite
imagery and other sensor-derived measurements, through masked reconstruction,
contrastive invariance, and cross-modal alignment objectives~\cite{cong2022satmae,manas2021seasonal,liu2024remoteclip}.
The resulting representation $\mathbf{s}_i$ encodes physical morphology such
as texture, spectral response, built density, and vegetation, but not the
functional role a region plays within the wider urban system.

\textbf{Region Representation Learning.} Given a set of regions $\mathcal{R}$ and their associated features, region representation learning aims to train an encoder $f: \mathcal{R} \rightarrow \mathbb{R}^d$ that maps each region $r_i \in \mathcal{R}$ to a low-dimensional embedding $\mathbf{z}_i \in \mathbb{R}^d$ that captures the region $r_i$'s characteristics and can be directly applied to a variety of downstream urban prediction tasks with a lightweight model such as Ridge Regression \cite{hoerl1970ridge}.

\subsection{Problem Statement}

We consider the problem of augmenting pretrained geospatial foundation
representations with the city-wide functional structure encoded by
human mobility. Given source cities
\(\mathcal{C}_{\mathrm{src}}\) with fixed foundation embeddings and
observed mobility graphs, the objective is to learn a region
representation model that transfers zero-shot to unseen target cities
\(\mathcal{C}_{\mathrm{tgt}}\), without updating the underlying
foundation model or performing target-city training; zero-shot thus
refers to the absence of target-city parameter updates, not the absence
of target-city inputs. At inference time,
the target city's mobility graph may be available or unavailable---in the
latter setting, we use a student model that only requires pretrained foundation
embeddings and inter-region geographic information.

\section{Methodology}\label{sec:model}

\subsection{Framework Overview}

MoRAX extends pretrained geospatial foundation representations with
the city-wide functional structure revealed by human mobility. Its
central operation is a lightweight mobility-induced modulation that
uses relational context to reinterpret the pretrained representation
without discarding its original geospatial semantics. 

When an observed mobility graph is available, a mobility-aware teacher
derives the conditioning context directly from inter-region flows.
When mobility is unavailable, a graph-free student approximates the
teacher-induced modulation using only pretrained geospatial embeddings
and inter-region geographic information.

\begin{figure*}
    \centering
      \includegraphics[
    width=\linewidth,
    trim=10 30 10 20,
    clip
]{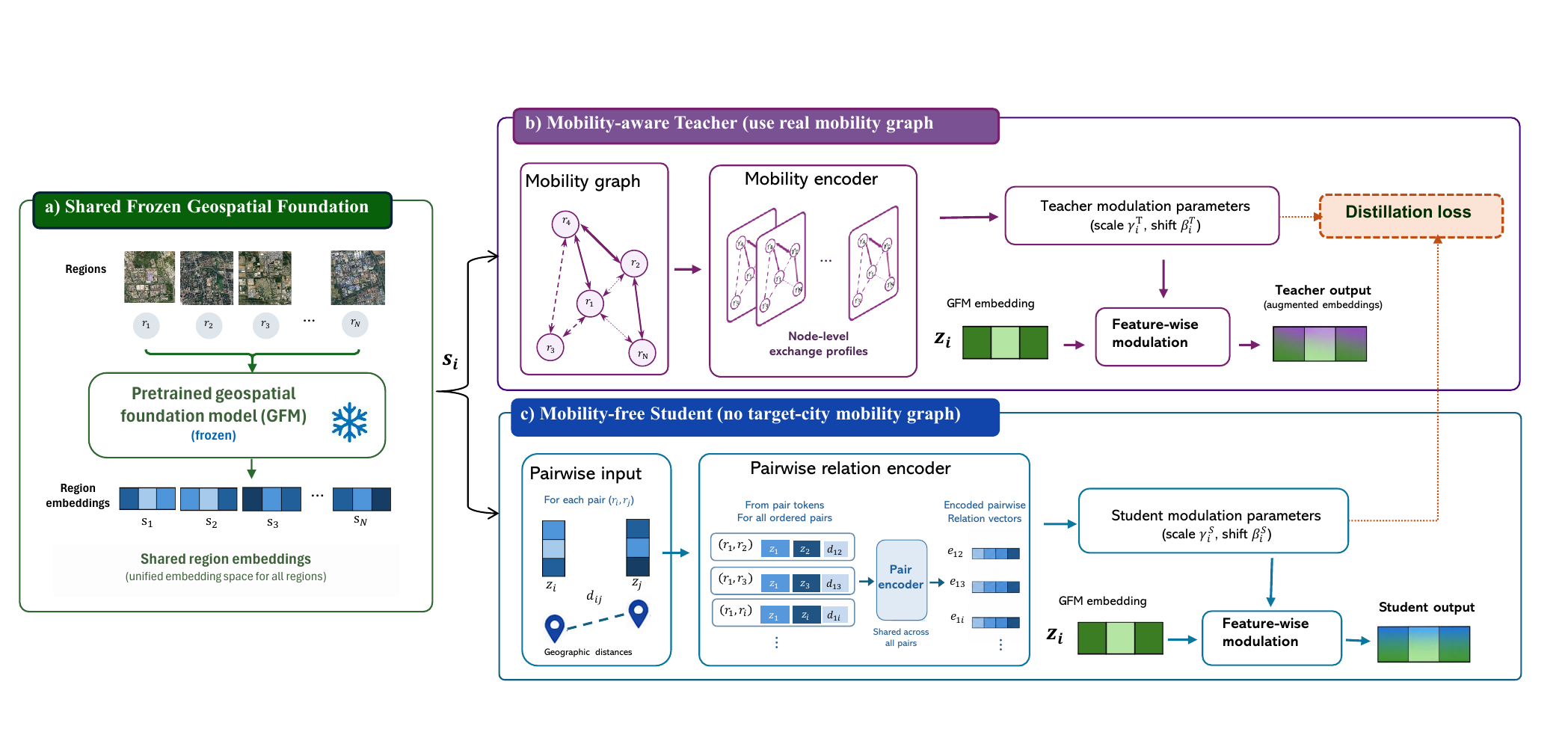}

    \caption{\textbf{The MoRAX architecture.} \normalfont (a)~A frozen geospatial foundation model provides a shared semantic base in which pretrained region embeddings are projected into a common representation space used by both teacher and student. (b)~The mobility-aware teacher encodes the observed mobility graph with an edge-centric message passing encoder and generates feature-wise scale and shift parameters that modulate the shared base. (c)~The mobility-free student approximates this modulation from pairwise geospatial embeddings and inter-region distances alone, trained by distilling the teacher's mobility-induced modulation.}
    \label{fig:model}
\end{figure*}

As illustrated in Fig. \ref{fig:model}, MoRAX first projects the pretrained GFM embedding of each region into a shared urban representation space
(Sec.~\ref{sec:geospatial_field}).  It then
defines a general feature-wise modulation operator that adapts the projected 
foundation representation according to a region-level relational
context (Sec.~\ref{sec:modulation}). The teacher obtains this context
from observed mobility relations through an edge-centric mobility
encoder (Sec.~\ref{sec:teacher}), whereas the student infers a proxy
context from pairwise geospatial representations
(Sec.~\ref{sec:student}). Finally, modulation distillation transfers
the mobility-induced adaptation from the teacher to the
student
(Sec.~\ref{sec:training}).

\subsection{Geospatial Foundation Representation}
\label{sec:geospatial_field}

As shown in Fig.~\ref{fig:model}(a), for each region $r_i$, we retrieve a pretrained GFM embedding
$\mathbf{s}_i \in \mathbb{R}^{d_{\mathrm{EO}}}$ according to its geographic location and spatial extent. These embeddings summarize EO-derived characteristics in a globally consistent representation space shared across source and target cities.

We map each pretrained embedding into the hidden
space of MoRAX using a lightweight projector
\begin{equation}
    \mathbf{z}_i^0
    =
    f_{\mathrm{proj}}(\mathbf{s}_i)
    \in \mathbb{R}^{d},
    \label{eq:geospatial_field}
\end{equation}
where $f_{\mathrm{proj}}$ is implemented as an MLP shared across all regions and cities. The underlying geospatial foundation representation remains
frozen and the projector is shared between the student and the teacher.

The resulting set
\(\mathbf{Z}^{0}=\{\mathbf{z}_i^0\}_{i=1}^{N}\) provides the shared
geospatial foundation representation used by both the teacher and student.
Because it is independent of the mobility graph, it provides a shared
semantic coordinate system across cities, regardless of whether mobility data are available. We refer to this spatially indexed layer of
region embeddings as
the \textit{geospatial field}. This field captures the
observable characteristics of individual regions, but it does not explicitly
encode how they interact within the boarder urban system. We therefore conditionally reshape this representation using
region-level relational context.

\subsection{Relation-Induced Feature Modulation}
\label{sec:modulation}

We first define the modulation operator, which is independent of how the conditioning signal is obtained. Given the projected foundation embedding $\mathbf{z}_i^0$ and a region-level relational context $\mathbf{z}_i^r$, a lightweight modulation network generates
feature-wise scale and shift parameters
\begin{equation}
    \boldsymbol{\gamma}_i
    =
    f_{\gamma,\theta}(\mathbf{z}_i^r),
    \qquad
    \boldsymbol{\beta}_i
    =
    f_{\beta,\theta}(\mathbf{z}_i^r),
    \label{eq:modulation_parameters}
\end{equation}
where
\(\boldsymbol{\gamma}_i,\boldsymbol{\beta}_i
\in\mathbb{R}^{d}\), and \(\theta\) denotes the parameters of the
modulation module.

We parameterize the feature-wise scale as
$1+\tanh(\boldsymbol{\gamma}_i)$, which constrains each scaling
coefficient to \((0,2)\). This parameterization centers the scale at identity and allows mobility to induce a bounded, feature-wise reweighting of the projected base. The resulting relation-conditioned
transformation is
\begin{equation}
    \mathbf{z}_i^{\mathrm{mod}}
    =
    \left(
        1+\tanh(\boldsymbol{\gamma}_i)
    \right)
    \odot
    \mathbf{z}_i^0
    +
    \boldsymbol{\beta}_i,
    \label{eq:modulated_features}
\end{equation}
where \(\odot\) denotes element-wise multiplication.

For compactness, we denote the complete operation in
Eqs.~\eqref{eq:modulation_parameters} and
\eqref{eq:modulated_features} as
\begin{equation}
    \mathbf{z}_i^{\mathrm{mod}}
    =
    \mathcal{M}_{\theta}
    \left(
        \mathbf{z}_i^0,
        \mathbf{z}_i^r
    \right).
    \label{eq:modulation_operator}
\end{equation}

The modulated features are then integrated with the projected
foundation representation
\begin{equation}
    \mathbf{z}_i
    =
    \mathbf{z}_i^0
    +
    \mathbf{z}_i^{\mathrm{mod}}.
    \label{eq:relation_adapted_representation}
\end{equation}

This additive formulation retains the projected geospatial
representation while allowing relational context to alter its
feature-wise interpretation. The projected foundation representation continues to provide the semantic basis of the final representation,
while the modulation introduces complementary information about the
functional role of each region.

The modulation operation is agnostic to the source of
\(\mathbf{z}_i^r\). In the mobility-aware teacher, the relational
context is encoded directly from observed mobility flows. In the
graph-free student, it is inferred from pairwise geospatial
representations and inter-region geography. The teacher and student
use the same modulation form but have independently learned parameters,
denoted by \(\theta_T\) and \(\theta_S\), respectively, with only the
frozen foundation embeddings and the projector \(f_{\mathrm{proj}}\) being shared between the two.

\subsection{Teacher: Mobility-Aware Relational Encoding}\label{sec:teacher}

When inter-region mobility flows are observed, we derive the relational
context directly from the mobility graph. We refer to this mobility-aware instantiation as the teacher because its induced
modulation later supervises the graph-free student.

Because mobility is naturally defined over region
pairs rather than individual regions, we adopt an edge-centric encoder
that explicitly updates relation states before aggregating them into
region-level mobility contexts, as shown in Fig.~\ref{fig:model}(b).

\subsubsection{Edge-Centric Mobility Encoding}
\label{sec:mobility_encoder}

Human mobility is naturally observed on region pairs rather than on individual regions. For each mobility edge \((r_i,r_j)\in\mathcal{E}_m\), let \(\mathbf{a}_{ij}^m\) denote its edge attributes, including within-city standardized flow strength and geographic distance. We first map these attributes into a hidden relation state
\begin{equation}
\mathbf{e}_{ij}^{(0)}
=
f_{\mathrm{edge}}(\mathbf{a}_{ij}^m).
\label{eq:mob_edge_init}
\end{equation}

The mobility encoder then applies \(L\) edge-centric message-passing layers. At layer \(\ell\), each region summarizes its incident relation states into a relational profile
\begin{equation}
\mathbf{h}_{i}^{(\ell)}
=
\operatorname{Agg}
\left(
\left\{
\mathbf{e}_{ji}^{(\ell-1)}
:
(r_j,r_i)\in\mathcal{E}_m
\right\}
\right),
\label{eq:mobility_profile}
\end{equation}
where \(\operatorname{Agg}\) is mean pooling in our implementation,
and \(\mathbf{h}_{i}^{(\ell)}\) is set to the zero vector for regions
with no incoming edges.

Each edge state is then updated using the relational profiles of both endpoints and its previous state
\begin{equation}
\mathbf{e}_{ij}^{(\ell)}
=
\mathbf{e}_{ij}^{(\ell-1)}
+
f_{\mathrm{upd}}^{(\ell)}
\left(
\mathbf{h}_{i}^{(\ell)}
\Vert
\mathbf{h}_{j}^{(\ell)}
\Vert
\mathbf{e}_{ij}^{(\ell-1)}
\right),
\label{eq:mobility_edge_update}
\end{equation}
where \(\Vert\) denotes vector concatenation and \(f_{\mathrm{upd}}^{(\ell)}\) is a layer-specific MLP. Since the readout below consumes only the relational profiles, the edge update is omitted at the final layer \(\ell=L\).

Finally, the mobility context of region \(r_i\) is read out from the relational profiles across all layers
\begin{equation}
\mathbf{z}_i^m
=
f_{\mathrm{read}}
\left(
\mathbf{h}_i^{(1)}
\Vert
\cdots
\Vert
\mathbf{h}_i^{(L)}
\right)
\in\mathbb{R}^{d}.
\label{eq:mob_readout}
\end{equation}

Since \(\mathbf{z}_i^{m}\) is constructed from explicitly updated relation
states, it summarizes the multi-layer relational profile
associated with region \(r_i\), rather than merely smoothing
neighboring node features. The resulting mobility context is then used as the relational signal for the teacher, \(\mathbf{z}_i^{r,T}=\mathbf{z}_i^{m}\).

The teacher then applies the modulation operator defined in
Eq.~\eqref{eq:modulation_operator}
\begin{equation}
    \mathbf{z}_i^{\mathrm{mod},T}
    =
    \mathcal{M}_{\theta_T}
    \left(
        \mathbf{z}_i^0,
        \mathbf{z}_i^{r,T}
    \right),
    \qquad
    \mathbf{z}_i^T
    =
    \mathbf{z}_i^0
    +
    \mathbf{z}_i^{\mathrm{mod},T}.
    \label{eq:teacher_adaptation}
\end{equation}

The teacher therefore uses observed city-wide interactions to reinterpret
the foundation representation while preserving its original geospatial
content. However, this requires access to the target city's
mobility graph. To remove the dependency at inference time, We introduce a graph-free student that
predicts the teacher's mobility-induced modulation from signals that are available even when mobility observarions are absent.

\subsection{Student: Graph-Free Modulation Approximation}
\label{sec:student}

As shown in Fig.~\ref{fig:model}(c), the student does not attempt to reconstruct the target mobility graph.
Instead, it learns to approximate how the mobility-aware teacher
reshapes each region representation.  It operates on the same
projected foundation representation as the teacher but infers
its relational context from pairwise geospatial embeddings and
inter-region geographic information.

\subsubsection{Geospatial Pair Relation Encoder}

For an anchor region $r_i$ and
a candidate partner $r_j$, where $j \neq i$, we construct the pair
attribute
\begin{equation}
    \mathbf{a}_{ij}^{p}
    =
    \left[
        \mathbf{s}_{i}
        \,\Vert\,
        \mathbf{s}_{j}
        \,\Vert\,
        d_{ij}
    \right],
    \label{eq:student_pair_attribute}
\end{equation}
where $\mathbf{s}_{i}$ and $\mathbf{s}_{j}$ are the pretrained
GFM embeddings of the two regions, and $d_{ij}$ is their geographic distance standardized
within the city.

A shared pair encoder maps each pair attribute into a
latent relation representation
\begin{equation}
    \mathbf{e}_{ij}^{S}
    =
    f_{\mathrm{pair}}
    \left(
        \mathbf{a}_{ij}^{p}
    \right)
    \in\mathbb{R}^{d}.
    \label{eq:student_pair_encoding}
\end{equation}

For each anchor region, the student aggregates the relation
representations of all candidate partners
\begin{equation}
    \mathbf{z}_{i}^{p}
    =
    \frac{1}{N-1}
    \sum_{j \neq i}
    \mathbf{e}_{ij}^{S},
    \label{eq:student_relation_context}
\end{equation}

The pair encoder captures the geospatial relationship between the anchor region and each candidate region, and mean pooling produces a
permutation-invariant summary over the city. This aggregated
representation serves as the relational conditioning
signal for the student, \(\mathbf{z}_i^{r,S}=\mathbf{z}_i^{p}\).

The student applies the same modulation form with independently learned
parameters
\begin{equation}
    \mathbf{z}_i^{\mathrm{mod},S}
    =
    \mathcal{M}_{\theta_S}
    \left(
        \mathbf{z}_i^0,
        \mathbf{z}_i^{r,S}
    \right),
    \qquad
    \mathbf{z}_i^S
    =
    \mathbf{z}_i^0
    +
    \mathbf{z}_i^{\mathrm{mod},S}.
    \label{eq:student_adaptation}
\end{equation}

\subsection{Training Objectives}\label{sec:training}

We train MoRAX in two stages. The teacher is first optimized on
source-city mobility graphs so that observed interaction structure is
encoded into both the mobility context and the final adapted
representation. The teacher is then frozen, and its mobility-induced
modulation supervises the graph-free student.

\subsubsection{Teacher Training}
The teacher is trained using self-supervision derived from observed
mobility relations. The objective is applied both to the mobility
context and to the final teacher representation.

\textbf{Mobility Relation Loss.}
For each anchor region \(r_i\), we construct a candidate set
\(\mathcal{C}_i=\mathcal{P}_i\cup\mathcal{N}_i\), where
\(\mathcal{P}_i\) contains all regions with observed mobility interactions
from \(r_i\), and \(\mathcal{N}_i\) contains distance-stratified sampled
non-neighbors. 

Positive candidates are assigned flow-normalized soft targets
\begin{equation}
    q_{ij}
    =
    \frac{w_{ij}}
    {\sum_{r_k\in\mathcal{P}_i}
    w_{ik}},
    \qquad
    r_j\in\mathcal{P}_i,
    \label{eq:mobility_soft_target_positive}
\end{equation}
while negative candidates receive zero target mass \(
    q_{ij}=0.
   \)
Given a region representation \(\mathbf{u}_i\), a level-specific pair scorer
\(\psi_u\) predicts a logit for each candidate
\begin{equation}
    s_{ij}^{u}
    =
    \psi_u
    \left(
        [
        \mathbf{u}_i
        \Vert
        \mathbf{u}_j
        ]
    \right),
    \qquad
    p_{ij}^{u}
    =
    \frac{\exp(s_{ij}^{u})}
    {\sum_{r_k\in\mathcal{C}_i}
    \exp(s_{ik}^{u})}.
    \label{eq:mobility_relation_prediction}
\end{equation}
The mobility relation loss is the soft-target cross entropy
\begin{equation}
    \mathcal{L}_{\mathrm{mrel}}(\mathbf{u})
    =
    -
    \frac{1}{|\mathcal{A}|}
    \sum_{r_i\in\mathcal{A}}
    \sum_{r_j\in\mathcal{C}_i}
    q_{ij}
    \log p_{ij}^{u},
    \label{eq:mobility_relation_loss}
\end{equation}
where \(\mathcal{A}\) denotes anchors with at least one observed
positive interaction.

We apply this objective at two levels. First, 
\(\mathcal{L}_{\mathrm{mrel}}^m=
\mathcal{L}_{\mathrm{mrel}}(\mathbf{z}^m)\) directly supervises the mobility encoder. Second,
\(\mathcal{L}_{\mathrm{mrel}}^T=
\mathcal{L}_{\mathrm{mrel}}(g_T(\mathbf{z}^T))\) applies the same supervision
to a relation-head projection of the final teacher representation, where \(g_T\) is a lightweight MLP head, to preserve mobility-derived interaction structure.

The overall teacher objective is then as follows
\begin{equation}
\mathcal{L}_T
=
\mathcal{L}_{\mathrm{mrel}}^m
+
\lambda_T\mathcal{L}_{\mathrm{mrel}}^T,
\label{eq:teacher_objective}
\end{equation}
where \(\lambda_T\) controls the contribution of the
representation-level relation objective. Note that the teacher
objective is purely relational, since the underlying geospatial semantics
are preserved architecturally, through the frozen foundation
embeddings and the additive formulation of
Eq.~\eqref{eq:relation_adapted_representation}, rather than through an
explicit grounding term as in the student objective below.

\subsubsection{Student Training} 
The student
learns to approximate the mobility-conditioned modulation generated by
the teacher.

\textbf{Distillation of relation-induced modulation.} We align the student modulated features with the detached teacher
features
\begin{equation}
    \mathcal{L}_{\mathrm{mod}}
    =
    \frac{1}{N}
    \sum_{i=1}^{N}
    \left\|
        \mathbf{z}_i^{\mathrm{mod},S}
        -
        \mathbf{z}_i^{\mathrm{mod},T}
    \right\|_2^2.
    \label{eq:modulation_distillation}
\end{equation}
To keep the student representation anchored to the pretrained
GFM semantic space, we use a lightweight decoder
\(\operatorname{Dec}_{S}\) to reconstruct the original foundation
embedding
\begin{equation}
    \mathcal{L}_{\mathrm{field}}^{S}
    =
    \frac{1}{N}
    \sum_{i=1}^{N}
    \left\|
        \operatorname{Dec}_{S}
        \left(
            \mathbf{z}_i^{S}
        \right)
        -
        \mathbf{s}_i
    \right\|_2^2.
    \label{eq:student_field_grounding}
\end{equation}
The overall student objective is
\begin{equation}
    \mathcal{L}_{S}
    =
    \alpha_{\mathrm{mod}}
    \mathcal{L}_{\mathrm{mod}}
    +
    \mathcal{L}_{\mathrm{field}}^{S},
    \label{eq:student_objective}
\end{equation}
where \(\alpha_{\mathrm{mod}}\) controls the weight of the modulation
distillation term.
\section{Experiments}

\begin{table*}[t]
\centering
\caption{Comparison of model performance on cross-city and cross-country transfer.
\normalfont Results are reported as $R^2$. The best result for each task is shown in bold and the second best is underlined. $\Delta$ denotes the relative improvement over the strongest non-MoRAX baseline for each task. Standard deviations over 10 random seeds are reported in parentheses (RemoteCLIP and AlphaEarth embeddings are deterministic and therefore reported without variance).}
\label{tab:main_results}
\scriptsize
\setlength{\tabcolsep}{3.5pt}
\renewcommand{\arraystretch}{0.98}
\resizebox{\textwidth}{!}{%
\begin{tabular}{lcccc|cccc}
\toprule
\multirow{2}{*}{Model} &
\multicolumn{4}{c|}{Shanghai} &
\multicolumn{4}{c}{Guangzhou} \\
\cmidrule(lr){2-5}\cmidrule(lr){6-9}
& Crime & NTL & CO$_2$ & PM$_{2.5}$ & Crime & NTL & CO$_2$ & PM$_{2.5}$ \\
\midrule
RemoteCLIP & 0.494 & 0.636 & 0.449 & 0.397 & 0.524 & 0.611 & 0.433 & 0.346 \\
AlphaEarth & 0.548 & 0.638 & 0.508 & 0.506 & 0.672 & 0.694 & 0.466 & 0.430 \\
\midrule
FlexiReg  & 0.192{\scriptsize(0.037)} & 0.237{\scriptsize(0.040)} & 0.228{\scriptsize(0.038)} & 0.281{\scriptsize(0.050)} & 0.209{\scriptsize(0.052)} & 0.225{\scriptsize(0.060)} & 0.257{\scriptsize(0.058)} & 0.399{\scriptsize(0.079)} \\
HREP & 0.268{\scriptsize(0.014)} & 0.361{\scriptsize(0.027)} & 0.323{\scriptsize(0.006)} & 0.592{\scriptsize(0.044)} & 0.367{\scriptsize(0.025)} & 0.415{\scriptsize(0.036)} & 0.355{\scriptsize(0.027)} & 0.640{\scriptsize(0.035)} \\
UrbanCLIP & 0.519{\scriptsize(0.007)} & 0.642{\scriptsize(0.004)} & 0.418{\scriptsize(0.001)} & 0.287{\scriptsize(0.017)} & 0.563{\scriptsize(0.008)} & 0.633{\scriptsize(0.008)} & 0.369{\scriptsize(0.024)} & 0.281{\scriptsize(0.019)} \\
\midrule
MoRAX-Student & \underline{0.569{\scriptsize(0.008)}} & \underline{0.774{\scriptsize(0.002)}} & \textbf{0.659{\scriptsize(0.002)}} & \underline{0.915{\scriptsize(0.001)}} & \underline{0.725{\scriptsize(0.017)}} & \underline{0.797{\scriptsize(0.018)}} & \textbf{0.702{\scriptsize(0.012)}} & \underline{0.929{\scriptsize(0.005)}} \\
Student $\Delta$ & $+3.8\%$ & $+20.6\%$ & $+29.7\%$ & $+54.6\%$ & $+7.9\%$ & $+14.8\%$ & $+50.6\%$ & $+45.2\%$ \\
MoRAX-Teacher & \textbf{0.618{\scriptsize(0.005)}} & \textbf{0.777{\scriptsize(0.010)}} & \underline{0.649{\scriptsize(0.004)}} & \textbf{0.917{\scriptsize(0.002)}} & \textbf{0.727{\scriptsize(0.005)}} & \textbf{0.803{\scriptsize(0.007)}} & \underline{0.696{\scriptsize(0.004)}} & \textbf{0.943{\scriptsize(0.002)}} \\
Teacher $\Delta$ & $+12.8\%$ & $+21.0\%$ & $+27.8\%$ & $+54.9\%$ & $+8.2\%$ & $+15.7\%$ & $+49.4\%$ & $+47.3\%$ \\
\midrule[\heavyrulewidth]
\multirow{2}{*}{Model} &
\multicolumn{4}{c|}{NYC} &
\multicolumn{4}{c}{Chicago} \\
\cmidrule(lr){2-5}\cmidrule(lr){6-9}
& Crime & Check-in & House & Income & Crime & Check-in & House & Income \\
\midrule
RemoteCLIP & 0.168 & 0.341 & 0.232 & 0.271 & 0.012 & 0.243 & 0.188 & 0.113 \\
AlphaEarth & 0.179 & 0.464 & 0.222 & 0.277 & 0.050 & 0.246 & 0.208 & 0.108 \\
\midrule
FlexiReg  & 0.176{\scriptsize(0.043)} & 0.256{\scriptsize(0.078)} & 0.179{\scriptsize(0.055)} & 0.209{\scriptsize(0.044)} & 0.174{\scriptsize(0.086)} & 0.233{\scriptsize(0.061)} & 0.307{\scriptsize(0.093)} & 0.235{\scriptsize(0.100)} \\
HREP & \underline{0.421{\scriptsize(0.039)}} & 0.564{\scriptsize(0.042)} & 0.419{\scriptsize(0.049)} & 0.518{\scriptsize(0.027)} & \underline{0.382{\scriptsize(0.052)}} & \underline{0.610{\scriptsize(0.053)}} & 0.625{\scriptsize(0.047)} & 0.455{\scriptsize(0.060)} \\
UrbanCLIP & 0.150{\scriptsize(0.033)} & 0.318{\scriptsize(0.038)} & 0.111{\scriptsize(0.018)} & 0.149{\scriptsize(0.026)} & 0.066{\scriptsize(0.021)} & 0.205{\scriptsize(0.054)} & 0.163{\scriptsize(0.048)} & 0.106{\scriptsize(0.037)} \\
\midrule
MoRAX-Student & 0.404{\scriptsize(0.013)} & \underline{0.588{\scriptsize(0.013)}} & \underline{0.466{\scriptsize(0.034)}} & \underline{0.530{\scriptsize(0.022)}} & 0.367{\scriptsize(0.030)} & 0.570{\scriptsize(0.040)} & \underline{0.715{\scriptsize(0.014)}} & \underline{0.502{\scriptsize(0.024)}} \\
Student $\Delta$ & $-4.0\%$ & $+4.3\%$ & $+11.2\%$ & $+2.3\%$ & $-3.9\%$ & $-6.6\%$ & $+14.4\%$ & $+10.3\%$ \\
MoRAX-Teacher & \textbf{0.520{\scriptsize(0.035)}} & \textbf{0.825{\scriptsize(0.020)}} & \textbf{0.469{\scriptsize(0.064)}} & \textbf{0.542{\scriptsize(0.039)}} & \textbf{0.662{\scriptsize(0.050)}} & \textbf{0.859{\scriptsize(0.021)}} & \textbf{0.809{\scriptsize(0.051)}} & \textbf{0.622{\scriptsize(0.051)}} \\
Teacher $\Delta$ & $+23.5\%$ & $+46.3\%$ & $+11.9\%$ & $+4.6\%$ & $+73.3\%$ & $+40.8\%$ & $+29.4\%$ & $+36.7\%$ \\
\bottomrule
\end{tabular}
}
\end{table*}

\subsection{Experimental Setup}

\textbf{Datasets.} We collect data from seven Chinese cities. Five cities, Beijing, Shenzhen, Hangzhou, Nanjing, and Suzhou, serve as source cities for pretraining, while \textbf{Shanghai (SH)} and \textbf{Guangzhou (GZ)} are held out for zero-shot transfer evaluation. We further include \textbf{New York City (NYC)} and \textbf{Chicago (CHI)} to assess cross-country transfer. We use AlphaEarth embeddings as the EO-based geospatial foundation representation in the main experiments. We adopt hexagonal H3 grids~\cite{h3_uber} at Level 7 for region discretization for Shanghai and Guangzhou, with each cell covering approximately $5.16~\mathrm{km}^2$, offering a spatial granularity suitable for city-level modeling. For NYC and CHI, administrative boundaries (i.e., census tracts) are adopted for region construction. We detail the training and downstream data collection and processing in Appendix \ref{app:datasets}.

\textbf{Implementation Details.}
MoRAX is trained in two stages. We first train the teacher on the 5 source cities with observed mobility graphs, and then freeze the teacher to train the student on the same source cities. Both stages use AdamW,
with learning rates \(3\times10^{-4}\) and \(1\times10^{-4}\) for the
teacher and student, respectively, and weight decay \(1\times10^{-2}\). The hidden dimension is \(d=128\), the mobility encoder uses \(L=3\) message-passing layers, and the loss weights are \(\lambda_T=0.5\) and \(\alpha_{\mathrm{mod}}=0.25\). For each anchor region, the negative set \(\mathcal{N}_i\) contains up to 24 non-neighbor regions
sampled with distance stratification. Specifically, candidate nagatives are divided into three distance-quantile bins according to their distance to \(r_i\),
and up to eight regions are sampled from each bin.

\textbf{Evaluation Procedure.} We evaluate two deployment regimes, corresponding to the teacher and student encoders introduced in Sec.~\ref{sec:model}. In the \textit{mobility-available} regime, the pretrained teacher $f_T$ is directly applied to a held-out target city using that city's own mobility graph, without any target-city fine-tuning. In the \textit{mobility-free} regime, the pretrained student $f_S$ is applied to the same held-out city using only widely available region signals, without observing the target city's mobility graph.

We consider four downstream prediction tasks for Shanghai and Guangzhou: crime, nighttime light intensity, carbon emissions, and PM$_{2.5}$ concentration. For New York City and Chicago, we evaluate crime, check-in counts, house prices, and income levels. For each downstream task, we train a lightweight Ridge regressor on the target city and evaluate it with five-fold cross-validation over regions. We report the coefficient of determination, $R^2$, as the main metric.

\textbf{Baselines.} 
We select state-of-the-art urban region representation methods that are
compatible with city-agnostic transfer as baselines, including
\textbf{HREP}~\cite{HREP}, \textbf{FlexiReg}~\cite{sun2025flexireg},
and \textbf{UrbanCLIP}~\cite{yan2024urbanclip}. For each method, we train
on the five source cities and directly apply the learned model to the
four test cities, reporting the average transfer performance across
source-city runs.

We further include two pretrained geospatial foundation models as
training-free baselines. \textbf{RemoteCLIP}~\cite{liu2024remoteclip}
directly encodes satellite imagery from the target cities, whereas
\textbf{AlphaEarth}~\cite{brown2025alphaearth} provides a pretrained
geospatial embedding for each target region, retrieved according to
its geographic location and spatial extent.

\subsection{Performance Comparison}
Table~\ref{tab:main_results} reports cross-city transfer results on four held-out cities, covering both within-country and cross-country settings. Overall, MoRAX-Teacher consistently outperforms pretrained foundation embeddings and existing urban representation baselines, achieving relative gains of up to \(54.9\%\) in the within-country setting and \(73.3\%\) in the cross-country setting.

Pretrained geospatial embeddings, particularly AlphaEarth, perform
strongly in the Chinese target cities, where tasks such as CO$_2$
prediction are closely associated with spatially continuous
physical and environmental patterns that Earth-observation-based
foundations can capture. Their advantage is less consistent in the
U.S. cities, where the prediction targets depend more strongly on human
activity and inter-region interactions and are less directly
recoverable from Earth-observation signals alone. This contrast
suggests that Earth-observation-based foundation
embeddings can benefit from complementary relational information.

Built on frozen AlphaEarth embeddings, MoRAX-Teacher increases the average \(R^2\) from \(0.558\) to \(0.766\) in the within-country setting and from \(0.219\) to \(0.663\) in the cross-country setting, corresponding to relative gains of \(37.4\%\) and \(202.6\%\), respectively. Importantly, these gains are not limited to tasks where the base GFM representation is weak. For example, AlphaEarth already performs well on NTL prediction in Shanghai and Guangzhou, yet relational adaptation still provides additional improvements. This suggests that mobility adds functional information that complements the foudnation representation, rather than simply correcting cases where the base embedding is insufficient. Overall, mobility conditioned modulation enables MoRAX to enrich geospatial foundation representations with interaction-aware information while preserving their observation-derived semantics.

MoRAX-Student further shows that a substantial portion of the
mobility-induced modulation can be transferred to a graph-free encoder
through distillation. On spatially smooth tasks, the student closely
matches the teacher and occasionally performs slightly better. In
contrast, the larger teacher--student gaps on crime and check-in
prediction suggest that some fine-grained human interaction patterns
remain difficult to recover without direct access to the target city's
mobility graph.

Nevertheless, MoRAX-Student improves over AlphaEarth by an average of
\(36.0\%\) in the within-country setting and \(136.1\%\) in the
cross-country setting. It also outperforms most urban region
representation baselines despite its lightweight, graph-free design.
The main exception is HREP on several interaction-intensive U.S. tasks,
where human mobility is used directly as an input.

\subsection{Cross-Country Transfer}
\label{sec:cross_country_transfer}

We further complement the cross-country evaluation with a bidirectional
transfer analysis. In addition to transferring models trained on the
five Chinese source cities to New York City and Chicago, we reverse the
direction by training MoRAX on the two U.S. cities and directly applying
it to the held-out Chinese cities.
\begin{table}[!htbp]
\centering
\caption{Cross-country transfer results ($R^2$).}
\label{tab:source_domain_transfer}
\setlength{\tabcolsep}{4pt}
\renewcommand{\arraystretch}{0.88}

\textbf{(a) Transfer to held-out Chinese cities}

\vspace{2pt}
\resizebox{\linewidth}{!}{%
\begin{tabular}{ll l cccc}
\toprule
\textbf{Target} & \textbf{Source} & \textbf{Encoder}
& \textbf{Crime} & \textbf{NTL} & \textbf{CO$_2$} & \textbf{PM$_{2.5}$} \\
\midrule
\multirow{5}{*}{SH}
& -- & AlphaEarth & 0.548 & 0.638 & 0.508 & 0.506 \\
& \multirow{2}{*}{5 CN} & Student & 0.569 & 0.774 & 0.659 & 0.915 \\
& & Teacher & 0.618 & 0.777 & 0.649 & 0.917 \\
& \multirow{2}{*}{NYC+CHI} & Student & 0.585 & 0.789 & 0.615 & 0.829 \\
& & Teacher & 0.625 & 0.785 & 0.597 & 0.824 \\
\midrule
\multirow{5}{*}{GZ}
& -- & AlphaEarth & 0.672 & 0.694 & 0.466 & 0.430 \\
& \multirow{2}{*}{5 CN} & Student & 0.725 & 0.797 & 0.702 & 0.929 \\
& & Teacher & 0.727 & 0.803 & 0.696 & 0.943 \\
& \multirow{2}{*}{NYC+CHI} & Student & 0.736 & 0.843 & 0.565 & 0.818 \\
& & Teacher & 0.742 & 0.849 & 0.541 & 0.782 \\
\bottomrule
\end{tabular}%
}

\vspace{6pt}

\textbf{(b) Transfer to U.S. cities}

\vspace{2pt}
\resizebox{\linewidth}{!}{%
\begin{tabular}{ll l cccc}
\toprule
\textbf{Target} & \textbf{Source} & \textbf{Encoder}
& \textbf{Crime} & \textbf{Check-in} & \textbf{House} & \textbf{Income} \\
\midrule
\multirow{5}{*}{NYC}
& -- & AlphaEarth & 0.179 & 0.464 & 0.222 & 0.277 \\
& \multirow{2}{*}{5 CN} & Student & 0.404 & 0.588 & 0.466 & 0.530 \\
& & Teacher & 0.520 & 0.825 & 0.469 & 0.542 \\
& \multirow{2}{*}{CHI} & Student & 0.349 & 0.537 & 0.402 & 0.536 \\
& & Teacher & 0.422 & 0.752 & 0.410 & 0.541 \\
\midrule
\multirow{5}{*}{CHI}
& -- & AlphaEarth & 0.050 & 0.246 & 0.208 & 0.108 \\
& \multirow{2}{*}{5 CN} & Student & 0.367 & 0.570 & 0.715 & 0.502 \\
& & Teacher & 0.662 & 0.859 & 0.809 & 0.622 \\
& \multirow{2}{*}{NYC} & Student & 0.453 & 0.517 & 0.651 & 0.425 \\
& & Teacher & 0.476 & 0.856 & 0.700 & 0.525 \\
\bottomrule
\end{tabular}%
}

\end{table}

As shown in Table~\ref{tab:source_domain_transfer}(a), training on the five
Chinese cities or on the two U.S. cities leads to broadly comparable
performance in Shanghai and Guangzhou. More importantly, both source configurations yield consistent
improvements over the unadapted AlphaEarth embeddings across all tasks
for both the student and teacher. This suggests that mobility-conditioned
modulation captures relational regularities that can transfer across
countries, rather than merely memorizing interaction patterns specific
to the source cities.

Interestingly, when transfering to New York City and Chicago, training on
the five Chinese cities generally performs better than training on the
other U.S. city (See Table \ref{tab:source_domain_transfer} (b)), despite the latter being geographically and
institutionally closer to the target. Averaged across the student and
teacher variants, using the five-city Chinese source set improves the mean
\(R^2\) from \(0.535\) to \(0.591\), corresponding to a relative gain of
\(10.4\%\), and achieves higher performance in 14 of the 16 model--task comparisons. This result suggests that source diversity may
be more important than geographic proximity. Exposure to mutiple training cities may help the model learn a modulation rule that is less tied to the regularities of any single city and therefore transfers more effectively.

\subsection{Generalization Across Geospatial Foundations}
Table~\ref{tab:backbone_comparison} evaluates whether MoRAX generalizes
beyond AlphaEarth by applying the same adaptation framework to
RemoteCLIP, a satellite-based vision-language foundation model. Across
Shanghai and Guangzhou, both MoRAX variants consistently improve the
original RemoteCLIP representations on all four downstream tasks.

To accommodate the independently encoded region images produced by
RemoteCLIP, we first transform them into a spatially continuous visual
field using parameter-free heat diffusion, as described in
Appendix~\ref{app:remoteclip_field}. We then apply the same relational
modulation framework to the geospatial field derived from RemoteCLIP.
The consistent gains across these two substantially different
foundation representations suggest that MoRAX is not tied to a particular geospatial backbone, but provides a general
interface for relational adaptation. Additional ablations in
Appendix~\ref{app:field_ablation} further show that constructing a
coherent semantic field is important for effective modulation.

\begin{table}
\centering
\caption{Performance applying MoRAX to RemoteCLIP ($R^2$). $\Delta$ denotes the relative improvement over RemoteCLIP.}
\label{tab:backbone_comparison}
\small
\setlength{\tabcolsep}{4pt}
\renewcommand{\arraystretch}{0.92}

\begin{tabular*}{0.98\columnwidth}{@{\extracolsep{\fill}}llcccc@{}}
\toprule
\textbf{City} & \textbf{Model}
& \textbf{Crime} & \textbf{NTL} & \textbf{CO$_2$} & \textbf{PM$_{2.5}$} \\
\midrule
\multirow{5}{*}{SH}
& RemoteCLIP       & 0.494 & 0.636 & 0.449 & 0.397 \\
& MoRAX-Student    & 0.532 & 0.716 & 0.632 & 0.874 \\
& Student $\Delta$ & +7.7\% & +12.6\% & +40.8\% & +120.2\% \\
& MoRAX-Teacher    & 0.596 & 0.739 & 0.623 & 0.884 \\
& Teacher $\Delta$ & +20.6\% & +16.2\% & +38.8\% & +122.7\% \\
\midrule
\multirow{5}{*}{GZ}
& RemoteCLIP       & 0.524 & 0.611 & 0.433 & 0.346 \\
& MoRAX-Student    & 0.655 & 0.748 & 0.687 & 0.895 \\
& Student $\Delta$ & +25.0\% & +22.4\% & +58.7\% & +158.7\% \\
& MoRAX-Teacher    & 0.702 & 0.801 & 0.635 & 0.931 \\
& Teacher $\Delta$ & +34.0\% & +31.1\% & +46.7\% & +169.1\% \\
\bottomrule
\end{tabular*}
\end{table}

\subsection{Ablation Study}
We conduct ablation studies on Shanghai to isolate the contribution of each component in MoRAX. For the teacher, \textbf{w/o modulation} directly uses the original AlphaEarth embedding without mobility-induced reshaping; \textbf{w/o relation} removes mobility-relation supervision and replaces it with geospatial reconstruction; \textbf{w/o foundation} removes the geospatial foundation and trains only the mobility encoder. For the student, \textbf{w/o distillation} trains the student directly from observed mobility relations without teacher supervision while \textbf{w/o modulation alignment} removes the objective aligning the student with the teacher's mobility-induced modulation.

\begin{figure}
    \centering
    \includegraphics[width=1\linewidth]{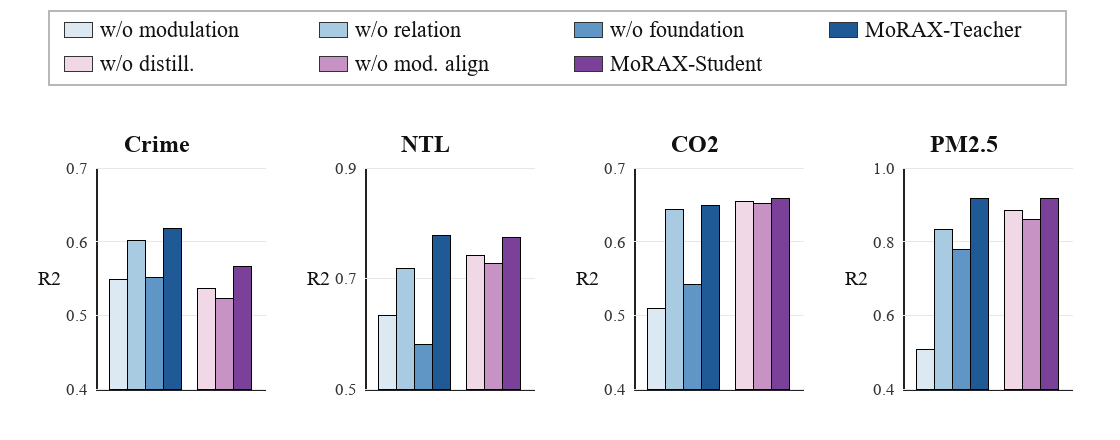}
    \caption{\textbf{Ablation study on Shanghai ($R^2$).} \normalfont Blue bars ablate the teacher by removing the modulation, the mobility-relation supervision, or the geospatial foundation. Purple bars ablate the student by training directly on mobility relations without the teacher (w/o distill.) or removing modulation alignment (w/o mod.\ align). The complete teacher and student match or exceed every ablated variant on all four tasks.}
    \label{fig:ablation_shanghai}
\end{figure}

As shown in Figure \ref{fig:ablation_shanghai}, the gap between the mobility-only variant (w/o foundation) and the teacher indicates that mobility provides useful relational information but benefits from the stable
semantic base of the geospatial foundation. Removing relation supervision also degrades the teacher, showing that reconstructing the original AlphaEarth embeddings alone
is insufficient to learn meaningful modulation. For the student, both
direct training and removing modulation alignment reduce performance on most tasks,
suggesting that teacher supervision helps transfer the
mobility-induced adaptation rather than only providing another
embedding target. 
Overall, the ablation results support the central design of MoRAX: a shared
geospatial foundation provides transferable semantics, mobility-relation
learning injects functional context through feature-wise modulation,
and distillation transfers this adaptation to the graph-free student.

\section{Conclusion}

We presented MoRAX, a framework that learns transferable urban region representations by using mobility to modulate a shared geospatial foundation. The key idea is that foundation embeddings provide a stable cross-city semantic base, but Earth-observation alone is not sufficient for urban understanding. MoRAX therefore treats mobility as relational context that reshapes the functional interpretation of geospatial foundation embeddings.

MoRAX further enables graph-free transfer by distilling the modulation process itself into a student that relies only on widely available proxy signals (e.g. pairwise distances). Across Chinese and U.S. cities, MoRAX consistently improves over pretrained geospatial embeddings and existing urban representation baselines. It further demonstrates the generalizability across foundation model backbones and across countries, strengthening the foundation backbone along a human-centric dimension. More broadly, this work suggests a shift in how relational signals can be incorporated into the growing ecosystem of pretrained foundation models.

\clearpage

\bibliographystyle{ACM-Reference-Format}
\bibliography{refs}

\appendix

\section{Constructing a Visual Field from RemoteCLIP} \label{app:remoteclip_field}

Unlike AlphaEarth, which directly provides a spatially contextualized
geospatial embedding field, RemoteCLIP independently encodes the
satellite image associated with each urban region. Directly applying
relational modulation to these isolated image embeddings provides no
explicit spatial continuity across neighboring regions. We therefore
apply a parameter-free heat diffusion operator before using RemoteCLIP
as the field base of MoRAX.

For each region $r_i$, a pretrained RemoteCLIP encoder produces an
image embedding
$\mathbf{s}_i\in\mathbb{R}^{d_{\mathrm{RC}}}$.
Let
$\mathbf{S}\in\mathbb{R}^{N\times d_{\mathrm{RC}}}$
denote the embedding matrix of all regions in a city. We construct a
distance-weighted $K$-nearest-neighbor graph ($K=8$) over region centroids.
The edge weight between neighboring regions is defined by an
exponential decay of geographic distance. The graph is then
symmetrized and row-normalized to obtain a transition matrix
$\mathbf{P}$.

Let $\mathcal{H}=\{0,1,2,4\}$ denote the retained diffusion orders.
The spatially diffused embedding of region $r_i$ is
\begin{equation}
    \widetilde{\mathbf{s}}_i
    =
    \sum_{h\in\mathcal{H}}
    w_h(t)
    \left(\mathbf{P}^{h}\mathbf{S}\right)_i,
    \label{eq:appendix_remoteclip_diffusion}
\end{equation}
where the diffusion weights follow a truncated and renormalized
Poisson distribution with diffusion time $t=2$:
\begin{equation}
    w_h(t)
    =
    \frac{
        e^{-t}t^h/h!
    }{
        \sum_{h'\in\mathcal{H}}
        e^{-t}t^{h'}/h'!
    }.
    \label{eq:appendix_remoteclip_weights}
\end{equation}

We then project the diffused satellite embedding into the model hidden space as
\(\mathbf{z}_i^0 = f_{\mathrm{proj}}(\widetilde{\mathbf{s}}_i) \in \mathbb{R}^{d}\). The resulting $\mathbf{z}_i^0$ serves as the RemoteCLIP-based visual
field for both the teacher and student. Heat diffusion is used only to
construct a spatially coherent field from independently encoded region
images, while the subsequent mobility encoder, pairwise student encoder,
modulation heads, and training objectives remain unchanged.

\section{Importance of the Semantic Field}
\label{app:field_ablation}

MoRAX assumes that relational context modulates a coherent region-level
semantic field. To examine the importance of this field structure, we
compare the complete RemoteCLIP-based model with a variant that applies
the same teacher and student modulation mechanisms directly to isolated
RemoteCLIP region embeddings without heat diffusion. All other
components and training settings are kept unchanged.

As reported in Table~\ref{tab:field_ablation}, removing heat diffusion consistently and largely weakens both the teacher and student. This result indicates that relational modulation benefits from an aligned and spatially coherent base: the mobility or proxy context can then reinterpret an existing semantic field rather than independently modulating a set of disconnected image embeddings.

\begin{table}[t]
\centering
\caption{Ablation on the RemoteCLIP-based semantic field.
``w/ Field'' applies heat diffusion before projection, whereas
``w/o Field'' directly projects isolated RemoteCLIP embeddings.
Results are reported in terms of $R^2$.}
\label{tab:field_ablation}

\small
\setlength{\tabcolsep}{3.5pt}
\renewcommand{\arraystretch}{0.95}

\begin{tabular*}{\columnwidth}
{@{\extracolsep{\fill}}llcc@{}}
\toprule
\multirow{2}{*}{\textbf{Task}}
& \multirow{2}{*}{\textbf{Variant}}
& \multicolumn{2}{c}{\textbf{Target Cities}} \\
\cmidrule(lr){3-4}
& & \textbf{GZ} & \textbf{SH} \\
\midrule

\multirow{4}{*}{Crime}
& Student w/o Field & 0.415  & 0.432 \\
& Student w/ Field  & 0.655  & 0.532 \\
& Teacher w/o Field & 0.571 & 0.534 \\
& Teacher w/ Field  & 0.702  & 0.596 \\
\midrule

\multirow{4}{*}{NTL}
& Student w/o Field & 0.572  & 0.628 \\
& Student w/ Field  & 0.748  & 0.716 \\
& Teacher w/o Field & 0.651 & 0.639 \\
& Teacher w/ Field  & 0.801  & 0.739 \\
\midrule

\multirow{4}{*}{CO$_2$}
& Student w/o Field & 0.237  & 0.449 \\
& Student w/ Field  & 0.687  & 0.632 \\
& Teacher w/o Field & 0.267  & 0.491 \\
& Teacher w/ Field  & 0.635 & 0.623 \\
\midrule

\multirow{4}{*}{PM$_{2.5}$}
& Student w/o Field & 0.345  & 0.384 \\
& Student w/ Field  & 0.895  & 0.874 \\
& Teacher w/o Field & 0.776  & 0.690\\
& Teacher w/ Field  & 0.931 & 0.884 \\
\bottomrule
\end{tabular*}
\end{table}

\section{Datasets}
\label{app:datasets}

\subsection{Training Datasets}
\textbf{Mobility graphs.} For U.S. cities, we use publicly available ride-hailing trip records released by local transportation agencies. New York City mobility flows are constructed from the Taxi and Limousine Commission (TLC) trip record data~\cite{nyctrip}, while Chicago mobility flows are constructed from the Transportation Network Providers (TNP) trip records released by the City of Chicago~\cite{chicago_transportation}. Both datasets provide origin--destination information, which we aggregate into region-level mobility graphs. For Chinese cities, mobility graphs are constructed by aggregating offline store-level transaction records from WeChat Pay (Tencent) into region-level interaction flows.

\textbf{AlphaEarth embeddings.} We use AlphaEarth embeddings as the geospatial foundation features in the main experiments. The embeddings are obtained from Google Earth Engine \cite{gorelick2017google}. For each H3 region, we extract all 10-meter AlphaEarth embedding pixels within the region boundary and use their mean vector as the region-level representation.

\subsection{Downstream Tasks}\label{app:downstream}

\textbf{Shanghai and Guangzhou.} For crime datasets, we aggregate the original crime cases data from \citet{crime} over a 5-year interval. 
Night-time light labels are derived by aggregating VIIRS radiance pixels within each H3 region. We remove invalid and negative pixels, compute region-level summary statistics, and use the log-transformed total radiance, \(\log(1+\mathrm{sum})\), as the downstream NTL target \cite{ntl}. Carbon emissions are derived from the ODIAC fossil-fuel CO2 emission dataset~\cite{CO2}. We use monthly 1 km ODIAC 2024 GeoTIFF rasters for 2021, aggregate pixel values within each H3 region across all months, and use the log-transformed annual sum.  PM2.5 concentration is derived from the CHAP / ChinaHighPM2.5 1 km annual PM2.5 product~\cite{pm2.5-1, pm2.5-2}. We use the 2021 NetCDF raster, aggregate valid raster pixels within each H3 region by their mean value.

\textbf{NYC and CHI.} Crime records are collected from NYPD \cite{nyc_opendata} and the Chicago Police Department~\cite{chicago_dataportal}. Check-in activity is obtained from Foursquare~\cite{Foursquare} and house price labels are derived from Zillow~\cite{zillow_research_data} and NYC Department of Finance records~\cite{nyc_finance}. We obtain regional income level data from the U.S. Census Bureau \cite{uscensus}.

\end{document}